\documentclass[conference]{IEEEtran}

\usepackage{cite}
\usepackage{amsmath,amssymb,amsfonts}
\usepackage{amsthm}

\usepackage{graphicx}
\usepackage{textcomp}
\usepackage{xcolor}
\usepackage{multirow}
\usepackage{booktabs}
\usepackage{soul}
\usepackage{array}
\usepackage{float}
\usepackage{tikz}
\usepackage{longtable}

\begin{document}

\title{Anomaly Detection via Federated Learning}

\author{Marc Vucovich$^{*}$$^{a}$,
        Amogh Tarcar$^{b}$,
        Penjo Rebelo$^{b}$,
        Narendra Gade$^{b}$,\\
        Ruchi Porwal$^{b}$
        Abdul Rahman$^{a}$,
        Christopher Redino$^{a}$,
        Kevin Choi$^{a}$,\\
        Dhruv Nandakumar$^{a}$,
        Robert Schiller$^{a}$,
        Edward Bowen$^{a}$,
        Alex West$^{a}$,\\
        Sanmitra Bhattacharya$^{a}$,
        Balaji Veeramani$^{a}$\\
        \small $^{a}$Deloitte \& Touche LLP \\
        \small $^{b}$Persistent Systems Limited\\
        \small $^{*}$Corresponding author: mvucovich@deloitte.com \\
}
\maketitle

\begin{abstract}
Machine learning has helped advance the field of anomaly detection by incorporating classifiers and autoencoders to decipher between normal and anomalous behavior. Additionally, federated learning has provided a way for a global model to be trained with multiple clients' data without requiring the client to directly share their data. This paper proposes a novel anomaly detector via federated learning to detect malicious network activity on a client’s server. In our experiments, we use an autoencoder with a classifier in a federated learning framework to determine if the network activity is benign or malicious. By using our novel min-max scalar and sampling technique, called \emph{FedSam}, we determined federated learning allows the global model to learn from each client's data and, in turn, provide a means for each client to improve their intrusion detection system’s defense against cyber-attacks.
\end{abstract}


\section{Introduction}
As cyber-attacks continue to evolve, the intrusion detection systems (IDS) used to detect these attacks need to stay up to date. To stay ahead of the attackers, companies should use the most advanced technology and work together in a way that allows them to contribute their insights in a secure manner. 

Our federated learning (FL) anomaly detection approach provides a method for a company to detect intrusion attacks while simultaneously contributing their insights to a global model without sharing their individual network activity. In a federated framework, the machine learning (ML) model is deployed and trained on individual systems, and the weights from the trained model are aggregated in a global model \cite{Rey}. In turn, the global model will continue to adjust based on the newly added weights and learn how to handle new data. When paired with network anomaly detection, companies can continuously share their insights and help the model learn to detect more attacks. 

Initial attempts at adopting FL in IDS have shown positive improvements. However, one of the key challenges while building an IDS is handling heterogeneous data distribution across multiple organizations \cite{sun}. In this paper, we will discuss some challenges that hamper interoperability of models across organizations and how FL can be used as a bridge to overcome these challenges. We propose training an autoencoder and classifier using our novel FL min-max algorithm and sampling technique called \emph{FedSam}.

The paper is structured as follows. First, we cover the background information for anomaly detection, autoencoders, and FL. Next, we discuss previous research related to our topic. Then, we describe the data sets used in our experiments, followed by the methods that lead to our anomaly detector via FL. After the methods, we discuss our experimental design and the results. We conclude with our final thoughts and our plans for future research.

\section{Background}

\subsection{Autoencoders and Anomaly Detection}

Autoencoders are one of the most popular neural network architectures for unsupervised anomaly detection \cite{alhajri}. A simple autoencoder comprises of an encoder block (one or more layer of neurons), a bottleneck (typically a layer with fewer neurons than the encoder) and a decoder block (same characteristics as the encoder), where the overall objective is to minimize the reconstruction loss when an input is passed through this network. Through this objective, the bottleneck layer of the autoencoder is able to capture the most representative features in a lower dimensional space. When an anomalous input is passed through an autoencoder trained on normal data, the reconstruction of the input is poor resulting in large reconstruction error. By establishing a threshold on the reconstruction error, an autoencoder can be used for the detection of anomalous inputs. In an IDS, attacks are sparse and benign traffic is abundant. Autoencoders can be trained to learn diverse benign traffic and minimize the average reconstruction loss \cite{Hindy}. Since the autoencoder has never encountered attack data during training, the reconstruction loss from attack data is typically higher than the reconstruction loss from benign data.
The threshold is the line that separates the two types of reconstruction losses such that the amount of benign data above the line is minimized and the amount of attack data above the line is maximized.

\subsection{Federated Learning and Anomaly Detection}

Traditionally, ML modeling techniques have relied on centralizing data from multiple sources into a single data center to train models. However, data about different types of intrusion attacks are rarely located at one organization. Often, attackers target multiple organizations, and the attack data is spread across them. Considering the sensitive nature of network related data, many organizations may find it challenging to share data for training ML models. Consequently, these organizations end up with an ML model that does not achieve its maximum potential. To resolve this issue, FL can be implemented. FL is a decentralized collaborative ML technique \cite{Rey}. Instead of aggregating data to create a single ML model, models are trained iteratively at every node, and the model parameters from each node are fused together using FL fusion algorithms \cite{Rey}.

FL is often implemented with a central FL server node orchestrating training rounds over multiple participating client nodes. At the beginning of each training round, the FL server shares a global FL model with each client node. Upon receiving the global FL model, each client runs ML training over the client's local data. These clients then send the updated model with learned parameters back to the FL server for aggregation. The FL server collects all the updates and fuses them by using one of the FL fusion algorithms. \emph{FedAvg} is one of the pioneering fusion algorithms \cite{campos}. Using \emph{FedAvg}, the global model update is obtained by the weighted average over all the parameters of each client model \cite{campos}. This completes one training round. Several training rounds are orchestrated by the FL server until the desired performance is achieved. This helps to ensure that client data never leaves its source location, and it allows multiple client nodes to collaborate and build a common ML model without directly sharing sensitive data.

\section{Literature Review}
With the constant advancements of ML techniques and the increased availability of intrusion detection data sets, researchers have been setting out to improve upon the current IDS. The variety of methods used to detect anomalies with ML have provided insights about the challenges of dealing with cyberattack data as well as possible solutions to overcome them.

The Canadian Institute for Cybersecurity 2017 Intrusion Detection System (CIC-IDS2017) and Canadian Institute for Cybersecurity 2018 Intrusion Detection System (CIC-IDS2018) data sets contain labeled network activity data for benign and malicious behavior \cite{Sharafaldin}\cite{brunswick}. Given the CIC-IDS data sets contain labeled data, a classification model is a logical approach to determine whether the data are benign or malicious. Zhou and Pezaros experimented with using 6 different types of classification models on the CIC-IDS2018 data set to determine if the data are ‘evil’ or ‘benign’ \cite{zhou}. The experiment initially tested each model on individual attacks, but in the final experiment the team used a decision tree classifier with each of the attack types grouped together as ‘evil’ data \cite{zhou}. The decision tree had an f-1 score of 1.0 detecting benign data and 0.57 detecting the attack data \cite{zhou}. The classifier had great results with detecting one type of attack, but it becomes increasingly difficult to differentiate between attacks as more types are added.

Although classification models have shown to be a viable approach, autoencoders have been very successful at detecting anomalies. Hindy \emph{et al.} conducted an experiment on the CIC-IDS2017 data set using an autoencoder with various threshold levels \cite{Hindy}. The autoencoder was trained using benign data so that the reconstruction loss would be higher when processing attack data. With the optimal threshold, the autoencoder had the following accuracies:90.01\%, 98.43\%, 98.47\%, and 99.67\% for DoS GoldenEye, DoS Hulk, Port Scanning, and DDoS attacks \cite{Hindy}. These results are very promising, but the varied accuracies based on the different threshold levels highlights the importance of using an optimal threshold. 

In another experiment, Li combines the autoencoder and classifier approaches to detect the attacks \cite{li}. To start the process, the normal data is sent through the autoencoder for dimensionality reduction \cite{li}. The data is then fed into a dense neural network that consists of 4 layers and an output layer for binary classification \cite{li}. The classifier's predictions were then used to train and test a decision tree \cite{li}. Along with Li’s experiment, Rezvy \emph{et al.} followed a similar approach using an autoencoder and a classifier \cite{rezvy}. The difference, however, is Rezvy \emph{et al.} use the autoencoder to minimize the reconstruction error \cite{rezvy}. The reconstruction error is then used as the input data for the classification model \cite{rezvy}. The results from this experiment are very promising, and the idea to use a classifier along with the autoencoder is a possible solution to finding the optimal threshold level.

In ``Chained Anomaly Detection Models for Federated Learning: An Intrusion Detection Case Study", Preuveneers \emph{et al.} built autoencoder based intrusion detection models using the CIC-IDS2017 data set \cite{preuveneers}. They partitioned data into 12 parties based on internet protocol (IP) addresses of victim machines. The autoencoders were trained using only benign traffic from the first day of CIC-IDS2017 simulations. In the experiments, the authors varied the number of parties from 1 to 12. 1 represented  the  central training and 12 represented the extreme case where each victim machine is a separate FL party \cite{preuveneers}. They observed that FL setups with more parties required more epochs for the model to converge. In their results, they claim that it took around 20 epochs for the central model to converge while the 12 party FL setup took around 50 epochs \cite{preuveneers}. While more epochs are needed, the amount of time for each epoch reduces as each party trains local models in parallel.  

In ``Federated Learning for Malware Detection in IoT Devices", Marmol Campos \emph{et al.} worked with malware detection using the N-BaIoT IOT dataset \cite{campos}. They described and compared two variations of \emph{FedAvg} algorithm: Mini-Batch Aggregation and Multi-Epoch Aggregation. In Mini-Batch Aggregation, data at party nodes are grouped into mini-batches for each FL round. Only a single mini-batch is used for training, and the updated model parameters are sent back to the FL server \cite{campos}. This process is repeated until all mini-batches are covered. In Multi-Epoch Aggregation, the received model is trained for multiple epochs using all the available data at a party node before sending model updates back to the server \cite{campos}. They described that an FL model trained with mini batch aggregation converges better than the multi-epoch aggregation \cite{campos}. One potential drawback of the multi-epoch aggregation approach is that model parameters in each FL round get optimized for a party node's local data as opposed to global training data. 

Another important contribution mentioned in paper ``Federated Learning for Malware Detection in IoT Devices" by Valerian Rey \emph{et al.}\cite{Rey} was calculating the min-max pre-processing scalar using a collaborative normalization algorithm. They suggested that training a common autoencoder with a global min-max is very important. They proposed an algorithm where each party node shares their min-max values with the FL server, and the FL server calculates the global min-max pre-processing scalar\cite{Rey}. They mentioned that this algorithm has a drawback of each client leaking exact values for each feature to the FL server, which creates a security threat to the client's data. In addition, as they used the autoencoder with a threshold as an anomaly detector, they needed to calculate the global threshold across multiple parties' data. They propose federating local thresholds, which requires each party to share their local threshold to the FL server where they are averaged as the global threshold.

Studies in FL have analyzed how the choice of FL algorithms affect training in IID (Identical and Independent) and Non-IID data distribution settings \cite{bahkti}. In Non-IID settings, mini-batch aggregation can lead to better convergence, however, it has a major drawback of being network in-efficient. In ``Federated Multi-Mini-Batch: An Efficient Training Approach to Federated Learning in Non-IID Environments", Bakhtiari \emph{et al.} suggested a new algorithm termed as \emph{FedMMB} (Federated Multi-Mini-Batch) \cite{bahkti}. This algorithm helps to overcome the challenge of network in-efficiency by training only on a subset of batches in every FL round. The number of batches used in every round is defined by a hyper-parameter, and these group of batches are run in an ordered sequence during successive FL rounds. Once all batches are completed, client data is shuffled and new batches are created \cite{bahkti}. This approach does help in aiding convergence, however, in case of dis-proportionate training data, this algorithm could allow clients to train for different amounts of epochs.

We used the insights and findings from each of these papers to help form our anomaly detector via FL. The research conducted before led to us researching ways to handle different data distributions, obtain a difficult threshold level, and account for different client sizes in FL.   

\section{Data Sets}
The CIC-IDS2017, CIC-IDS2018, National Collegiate Cyber Defense Competition (NCC-DC), and MAWI-Lab data sets were used to evaluate the anomaly detector via FL approach. The CIC-IDS2017 and CIC-IDS2018 datasets were created by the Canadian Institute for Cybersecurity (CIC), and they contain labeled network data for both benign and cyber-attacks \cite{Sharafaldin}\cite{brunswick}. The NCC-DC data set was created for the National Collegiate Cyber Defense Competition where Deloitte was one of the sponsors. It is important to note that only the sections the authors were involved with were used from the NCC-DC dataset \cite{williamsNCCDC}. Lastly, the MAWI lab data set is a database of benign activity \cite{mawilab}.

The CIC-IDS2017 dataset resembles real world network activity data (PCAPs) for benign and malicious behavior \cite{Sharafaldin}. It is broken into 5 csv files, and each file contains 80 features describing the activity as well as a label to indicate if the activity is benign or one of the following attacks: FTP-Patator, SSH-Patator, DoS Slowloris, DoS Slowhttptest, DoS Hulk, DoS GoldenEye, Heartbleed Port 444, Web Attack – Brute Force, Web Attack – XSS, Web Attack - SQL Injection, Infiltration, Exploit, Botnet, Port Scan, and DDoS LOIT  \cite{Sharafaldin}.

Similar to the CIC-IDS2017 data set, the CIC-IDS2018 data set is broken into 10 csv files \cite{brunswick}. Each file contains 80 features captured from the network traffic and system logs of 420 machines and 30 servers \cite{brunswick}. Additionally, the network traffic is labeled as either benign or as one of the following attacks: Brute-force FTP-Patator, Brute-force SSH-Patator, DoS Attack Hulk, DoS Attack Slowloris, DoS Attack GoldenEye, DoS Slowhttptest, DoS Attack Heartleech, Web Attack DVWA, Web Attack XSS, Web Attack Brute-force, Infiltration Attack, Botnet Attack, DDoS LOIC for UDP, DDoS LOIC for TCP, and DDoS LOIC for HTTP Requests \cite{brunswick}.

The next data set used was the section of the NCC-DC data set where the authors were responsible. It contains the same 80 features as the CIC-IDS data sets along with the attributes capturing the source port, and the source and destination IP addresses. The data set Deloitte had access to, had labeled network activity for benign and the following attack types: Scanning, Interrogation, Command and Control, and Exfiltration \cite{williamsNCCDC}.

The last data set used was from the MAWI-Lab in Japan. The data set contains 66 features, all of which are a subset of CIC-IDS/NCC-DC datasets, along with the attributes capturing the source port, source and destination IP addresses \cite{mawilab}. For evaluation purposes, we gathered a weeks’ worth of data from MAWI-Lab and processed it to fit the format of the CICIDS and NCC-DC datasets. The dataset contains 1,500,000 data points, and considering the dataset is a collection of real traffic events and attacks are very rare in the real world, we have assumed the data points are benign.

\section{Methods and Techniques}
This section of the paper will cover how the structure of the autoencoder and classifier as well the FL model architecture.

\begin{figure*}[ht!]
    \includegraphics[width=18cm]{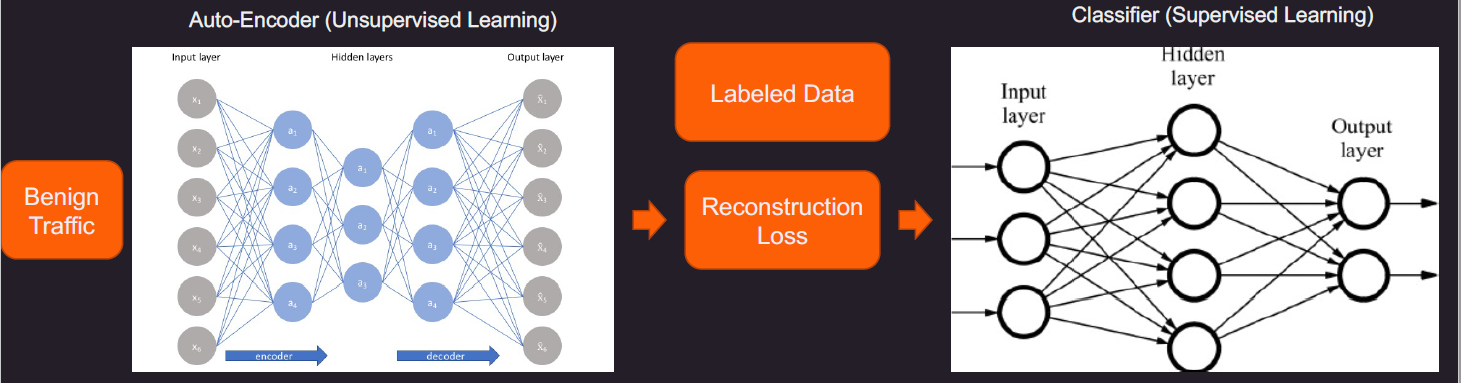}
    \centering
    \caption{Auto-Encoder and Classifier Block Diagram}
    \label{fig:1}
\end{figure*}

\subsection{Autoencoder and Classifier Setup}
The initial experiments were conducted using an undercomplete autoencoder with a root mean square propagation optimizer and mean squared error loss function. The autoencoders in these experiments were trained using benign data from the CIC-IDS2018 dataset and tested with benign and malicious data. 

As more attack types were passed through the autoencoder, it became increasingly difficult to determine a threshold that separated benign from malicious data. For this reason, a classifier was paired with the autoencoder to eliminate the need for a threshold \cite{li}\cite{rezvy}. The autoencoder was trained with benign data as mentioned in the previous paragraph. However, instead of setting a threshold level to classify the data, the reconstruction loss from the autoencoder was used to train a binary classification model. Using a balanced amount of benign and malicious data, the reconstruction loss produced from the autoencoder was used as the input data to train the sequential binary classifier. The classifier used a SoftMax activation function, Adam optimizer, and categorical cross entropy loss function. Additionally, the benign data was labeled as ‘0’ and all attack types were labeled as ‘1’ for training. The structure for the combined autoencoder and classifier is shown in figure 1.

Additionally, we trained the model with 75 features from the CIC-FlowMeter. Dst Port, Timestamp, Flow Byts/s, Flow Pkts/s and Labels were the only features from the CIC-IDS dataset that were left out of the training data. Adding extra features for training did not reduce the model performance, and it even helped detect more types of attacks. 

\subsection{Federated Learning Setup}
In our federated setup, we trained the autoencoder and classifier mentioned in the above section in a 3-step process. 

\begin{description}
    \item[1.] The FL server sent the autoencoder to each client to be trained with their own benign data. Each client then sent the weights back to the FL server to be aggregated.
    \item[2.] Each client used the global FL autoencoder to create training data for the classifier by running raw inputs over its local data and saving the reconstruction loss vectors for each data point. 
    \item[3.] The FL server sent the classifier model to each client and each client used their data from step 2 and their data labels to train the classifier. The weights from the trained model were then sent back to the FL server to be aggregated.
\end{description}

Throughout the process, only the model parameters are sent to the FL server. None of the actual data from the client leaves the client’s server. At the end of these three steps, we obtained a single global pair of an autoencoder and classifier.

Once the basic FL architecture was complete, we worked on handling clients with different data distributions. To do this, we introduced a novel global FL min-max scalar algorithm and a sampling technique we call \emph{FedSam}.

The novel min-max scalar works as follows. The FL server obtains a list of the IP addresses of the connected client nodes and initializes a min-max scalar object with random minimum and maximum values for each feature used to train the models. The FL server then randomly selects an IP address from the list and shares the initial min-max scalar with the client. That IP address is then deleted from the list. Once the client receives the min-max scalar, the client node will validate if its data is within the bounds of the min-max scalar. If it is within the bounds, the client node will not make any changes to the min-max scalar. If there are any values outside the bounds, the client node will update that specific feature with their value. When the value has been updated, the client node will randomly select another IP address from the list, send the min-max scalar to the next client and delete the IP address from the list. The next client will repeat the same exact steps as the previous client. This process will continue until every client has had a chance to update the min-max scalar. Once the process is complete, the common min-max scalar will have the minimum and maximum values for each feature. This min-max scalar will then be sent out by the FL server to each client to scale the data.

Our federated aggregation strategy - \emph{FedAvg} with sampling, called \emph{FedSam}, is a combination of Mini-Batch and Multi-Epoch \emph{FedAvg} strategy, and it is well suited for giving equal weight to updates from all client nodes. Multi-Epoch strategy averages model parameters in each round after training for one or more epochs of client data while single Mini-Batch strategy averages after one single mini-batch of client data and continues sequentially. With \emph{FedMMB} , they introduce a hyper-parameter which is count of mini-batches to train over in a single FL round and continue sequentially. In \emph{FedSam} we improve on \emph{FedMMB} and introduced a hyper parameter called sampling size replacing the batch count, and the sampling size is a number of data points which will be used for training in each round. Sample size is batch size multiplied by batch count in each round. We observed that in \emph{FedMMB}, there is some variation in the number of data points used in FL rounds when training with clients that have data imbalances. As its a weighted average, this might offset overall training of the FL model. Furthermore, as \emph{FedMMB} executes in sequence, extreme data imbalance will translate to some clients completing multiple epochs while others only complete one epoch. 

For example, with \emph{FedSam} if there are 100k rows of data, we could set the sampling size to be 5000 and batch size to be 20. This means, for 1 epoch, the 100k rows of data will be processed by the model in batches of 20 in each FL round. When used with 2 clients with different amounts of data, each client will train on the same amount of data in each FL round. As an example, if client 1 had 100k rows of data and client 2 had 30k rows of data, we would set the sample size to 5000 and batch size to 20. Each client would then sample 5000 data points from their dataset in each FL round. This method ensures every client uses the exact same amount of data to train their model and prevents one client from having too much influence on the global model over FL training rounds. 

\section{Experimental Design}
Once our anomaly detector via FL was solidified, we conducted multiple experiments to determine the effectiveness of the model. We have organized our experiments in manner that build on one another to highlight the usefulness of our final FL model. 

The experimental goals are listed below.

\begin{description}
    \item[1.] Create benchmark results using central models
    \item[2.] Demonstrate the effectiveness of our novel min-max scalar and sampling technique in an FL set up
    \item[3.] Compare \emph{FedSam} to \emph{FedMMB}
    \item[4.] Test our FL model on unknown data
\end{description}

For experiments A through C, we used the CIC-IDS2017, CIC-IDS2018, and NCC-DC datasets. The training and testing splits for each dataset are shown in table I. 

\begin{table}[ht!]
\centering
\renewcommand{\arraystretch}{1.5}
\begin{tabular}{ | m{5em} | m{7em}| m{6em} | m{5em} |} 
  \hline
  \textbf{Data} & \textbf{Autoencoder Training Split} & \textbf{Classifier Training Split} &  \textbf{Testing}
  \\
  \hline
  CICIDS2017 & Benign Ex: \newline 1136538 \newline\newline Attack Ex:\newline 0 & Benign Ex: \newline 61000 \newline\newline Attack Ex:\newline 60400 & Benign Ex: \newline 8000 \newline\newline Attack Ex:\newline 8000   \\
  \hline
  CICIDS2018 & Benign Ex: \newline 2279560 \newline\newline Attack Ex:\newline 0 & Benign Ex: \newline 23375 \newline\newline Attack Ex:\newline 23375 & Benign Ex: \newline 8000 \newline\newline Attack Ex:\newline 8000   \\
  \hline
  NCC-DC & Benign Ex: \newline 79848 \newline\newline Attack Ex:\newline 0 & Benign Ex: \newline 10000 \newline\newline Attack Ex:\newline 10000 & Benign Ex: \newline 8000 \newline\newline Attack Ex:\newline 8000   \\
  \hline
\end{tabular}
\label{table: 1}
\newline
\caption{Training and Testing splits for each dataset in experiments A-C}
\end{table}

\subsection{Central Models}
Our first set of experiments test how centrally trained autoencoders and classifiers fair against a global dataset. Three of the central models were trained individually with either the CIC-IDS2018, CIC-IDS2017, or NCC-DC dataset and tested with a combination of all three datasets. The results from these models are used to help highlight the advantage for clients to use FL. 

The fourth central model was trained with data from the CIC-IDS2018, CIC-IDS2017, and NCC-DC datasets, and it was tested using a combination of all three datasets. The results from this model serve as a theoretical benchmark for an all inclusive central model that we can use to compare to our federated model.

The structure of the models trained and tested is shown in the table II.

\begin{table}[ht!]
\centering
\renewcommand{\arraystretch}{1.5}
\begin{tabular}{ | m{6em} | m{10em}| m{8em} |} 
  \hline
  \textbf{Model} & \textbf{Training Data} & \textbf{Test Data} \\ 
  \hline
  1 & CICIDS2018 & NCC-DC \newline CIC-IDS2017 \newline CIC-IDS2018 \\ 
  \hline
  2 & CICIDS2017 & NCC-DC \newline CIC-IDS2017 \newline CIC-IDS2018   \\ 
  \hline
  3 & NCC-DC & NCC-DC \newline CIC-IDS2017 \newline CIC-IDS2018 \\
  \hline
  4 & NCC-DC \newline CIC-IDS2017 \newline CIC-IDS2018 & NCC-DC \newline CIC-IDS2017 \newline CIC-IDS2018 \\ 
  \hline
\end{tabular}
\label{table: 2}
\newline
\caption{Central Models}
\end{table}

\subsection{Novel Min-Max Scalar and FedSam}

Our next experiment was conducted to show that our min-max algorithm and sampling technique are effective methods to handle convergence issues when clients have different amounts of data. We trained a 3-party federated model using data from the NCC-DC, CIC-IDS2017, and CIC-IDS2018 data sets where each party used a min-max scalar relevant to their own data. Then, we trained another 3-party model with the same set up, but we used our novel min-max algorithm. Finally, we trained another 3-party model with the same set up, but we used our novel min-max algorithm with \emph{FedSam}. The results were used to show the effectiveness of our novel min-max scalar and sampling technique. 

\begin{table}[H]
\centering\
\renewcommand{\arraystretch}{1.5}
\begin{tabular}{ | m{6em} | m{10em}| m{8em}|} 
  \hline
   \textbf{Model} & \textbf{Clients} & \textbf{Technique}   \\ 
  \hline
  1 & Client 1: CIC-IDS2018 \newline Client 2: CIC-IDS2017 \newline Client 3: NCC-DC & Individual \newline MinMax Scalars\\ 
  \hline
  2 & Client 1: CIC-IDS2018 \newline Client 2: CIC-IDS2017 \newline Client 3: NCC-DC & Novel MinMax \newline Scalar \\ 
  \hline
  3 & Client 1: CIC-IDS2018 \newline Client 2: CIC-IDS2017 \newline Client 3: NCC-DC & Novel MinMax \newline Scalar and \newline \emph{FedSam}\\ 
  \hline
\end{tabular}
\label{table: 3}
\newline
\caption{Min-Max and \emph{FedSam} Experiments}
\end{table}

\subsection{FedSam vs. FedMMB}
Our next experiment compared the optimized FL model using our novel min-max algorithm and our \emph{FedSam} algorithm to an FL model using our novel min-max algorithm and the \emph{FedMMB} algorithm. We trained two separate federated models using 3 clients with the CIC-IDS2017, CIC-IDS2018, and NCC-DC datasets. We then used \emph{FedMMB} on the first federated model and \emph{FedSam} on the second federated model. 

\begin{table}[ht!]
\centering
\renewcommand{\arraystretch}{1.5}
\begin{tabular}{ | m{6em} | m{10em}| m{8em}|} 
  \hline
   \textbf{Model} & \textbf{Clients} & \textbf{Algorithm}   \\ 
  \hline
  1 & Client 1: NCC-DC \newline Client 2: CIC-IDS2017 \newline Client 3: CIC-IDS2018 & \emph{FedMMB}\\ 
  \hline
  2 & Client 1: NCC-DC \newline Client 2: CIC-IDS2017 \newline Client 3: CIC-IDS20188 & \emph{FedSam} \\ 
  \hline
\end{tabular}
\label{table: 4}
\newline
\caption{\emph{FedMMB} and \emph{FedSam} Experiments}
\end{table}
\subsection{Unknown Data}

Our final experiment tested our optimized FL model on unknown data sets. The models were trained with 3 clients using three of the following datasets: CIC-IDS2017, CIC-IDS2018, NCC-DC, or MAWI-Lab. The model was then tested with the entire dataset not used to train the model. The training splits for CIC-IDS2017, CIC-IDS2018, and NCC-DC are the same as shown in table I. MAWI used 59,138 benign examples to train the autoencoder. 

It is important to note the MAWI dataset does not have attack data, so it only participated in the FL for the autoencoder. \newline See table V.
\begin{table}[ht]
\centering
\renewcommand{\arraystretch}{1.5}
\begin{tabular}{ | m{3em} | m{11.5em}| m{12em}|} 
  \hline
   \textbf{Model} & \textbf{Clients} & \textbf{Test Data}   \\ 
  \hline
  1 & Client 1: CIC-IDS2018 \newline Client 2: CIC-IDS2017 \newline Client 3: NCC-DC & MAWI \newline \newline Benign Examples: 1,500,000 \newline Attack Examples: 0\\ 
  \hline
  2 & Client 1: Mawi-Lab \newline Client 2: CIC-IDS2017 \newline Client 3: NCC-DC & CIC-IDS2018 \newline \newline Benign Examples: 2,218,562 \newline Attack Examples: 1,623,490\\ 
  \hline
  3 & Client 1: CIC-IDS2018 \newline Client 2: MAWI-Lab \newline Client 3: NCC-DC & CIC-IDS2017 \newline \newline Benign Examples: 1,105,164 \newline Attack Examples: 526,203\\ 
  \hline
  4 & Client 1: CIC-IDS2018 \newline Client 2: CIC-IDS2017 \newline Client 3: MAWI-Lab & NCC-DC \newline \newline Benign Examples: 24,000 \newline Attack Examples: 431,870\\ 
  \hline
\end{tabular}
\label{table: 5}
\newline
\caption{FL Experiments with Unknown Data}
\end{table}

\section{Results}
 Our results show the novel min-max scalar algorithm, sampling technique, and FL method are effective solutions to build a secure and collaborative IDS. The results from the experiments are shown in the subsections below.
 
\subsection{Central Models}
The experiments with the central models are the benchmark performances to compare to our federated model. The individual model's results show that an individual IDS has a limited scope of attacks it can detect. However, as shown in the upcoming sub-sections, we can expand the scope by using a globally shared FL model. 
  
Additionally, the fourth model which was trained using all three datasets serves as a theoretical benchmark. These results could only be achieved if each client decided to store their data together on a central network. 
  
The results from the experiments are shown in table VI. They include precision, recall, f1 score, accuracy, and the confusion matrix for the following models. 
 
 \begin{description}
    \item[1.] Trained using CIC-IDS2018 data
    \item[2.] Trained using CIC-IDS2017 data
    \item[3.] Trained using NCC data
    \item[4.] Trained using all three data sets 
\end{description}

Each model was tested using a combination of benign and attack data from the CIC-IDS2017, CIC-IDS2018, and NCC-DC datasets shown in table I.

\begin{table}[ht!]
\centering
\renewcommand{\arraystretch}{1.5}
\begin{tabular}{ |m{2.5em}|m{2.75em}| m{2.75em}|m{2.75em}|m{3.5em}|m{5.5em}|} 
\hline
\textbf{Model} &
\textbf{Data} & \textbf{Pre-\newline cision} & \textbf{Recall} & \textbf{F1 \newline Score}  & \textbf{Confusion\newline Matrix}
\\\hline
1& Benign & 0.69 & 0.71 & 0.70 & [16985,7015]\\
\cline{2-5}
& Attack & 0.70 & 0.69 & 0.68 & [7509,16500]\\
\hline
2 & Benign & 0.66 & 0.53 & 0.59 & [12674,11326]\\
\cline{2-5}
& Attack & 0.61 & 0.73 & 0.66 & [6587,17413]\\
\hline
3 & Benign & 0.39 & 0.31 & 0.34 & [7352,16648]\\
\cline{2-5}
& Attack & 0.43 & 0.52 & 0.47  &[11506,12494\\
\hline
4 & Benign & 1.00 & 0.99 & 0.99 & [23831,169]\\
\cline{2-5}
& Attack & 0.99 & 1.00 & 0.99 & [119,23881]\\
\hline
\end{tabular}
\label{table: 6}
\newline
\caption{Centrally Trained Models}
\end{table}
 
 \subsection{Novel Min Max Scalar and FedSam}
 
 In these experiments we compared the results of an FL model using individual min-max scalars to an FL model using our novel min-max scalar and to an FL model using our novel min-max scalar with \emph{FedSam}. The results demonstrate the effectiveness of using our novel min-max algorithm and \emph{FedSam}. Additionally, these models were trained for 200 rounds since the models without \emph{FedSam} take much longer to train. Below are the clients used in each model. 
 
 \begin{description}
    \item[1.] Client 1 trained using CIC-IDS2018 data
    \item[2.] Client 2 trained using CIC-IDS2017 data
    \item[3.] Client 3 trained using NCC-DC data
\end{description}
 
\begin{table}[H]
\centering
\renewcommand{\arraystretch}{1.5}
\begin{tabular}{ |m{2.5em}|m{2.75em}| m{2.75em}|m{2.75em}|m{3.5em}|m{5.5em}|} 
\hline
\textbf{Client} &
\textbf{Data} & \textbf{Pre-\newline cision} & \textbf{Recall} & \textbf{F1 \newline Score}  & \textbf{Confusion\newline Matrix}

\\\hline
1 & Benign & 0.84 & 0.64 & 0.72 & [15330,8670]\\
\cline{2-5}
& Attack & 0.71 & 0.88 & 0.78 & [2977,21023]\\
\hline
2 & Benign & 0.65 & 0.70 & 0.67  & [16778,7222]\\
\cline{2-5}
& Attack & 0.67 & 0.62 & 0.64 & [9153,14847]\\
\hline
3 & Benign & 0.62 & 0.72 & 0.67 & [17327,6673]\\
\cline{2-5}
& Attack & 0.67 & 0.55 & 0.60 & [10729,13271]\\
\hline
\end{tabular}
\label{table: 7}
\newline
\caption{FL models with individual scalars}
\end{table}

\begin{table}[ht!]
\centering
\renewcommand{\arraystretch}{1.5}
\begin{tabular}{ |m{2.5em}|m{2.75em}| m{2.75em}|m{2.75em}|m{3.5em}|m{5.5em}|} 
\hline
\textbf{Model} &
\textbf{Data} & \textbf{Pre-\newline cision} & \textbf{Recall} & \textbf{F1 \newline Score}  & \textbf{Confusion\newline Matrix}

\\\hline
1 & Benign & 0.92 & 0.80 & 0.86  & [19138,4862]\\
\cline{2-5}
& Attack & 0.82 & 0.93 & 0.87 & [1574,22426]\\
\hline
\end{tabular}
\label{table: 8}
\newline
\caption{Federated Model results using our novel min-max scalar}
\end{table}

\begin{table}[ht!]
\centering
\renewcommand{\arraystretch}{1.5}
\begin{tabular}{ |m{2.5em}|m{2.75em}| m{2.75em}|m{2.75em}|m{3.5em}|m{5.5em}|} 
\hline
\textbf{Model} &
\textbf{Data} & \textbf{Pre-\newline cision} & \textbf{Recall} & \textbf{F1 \newline Score} & \textbf{Confusion\newline Matrix}

\\\hline
1 & Benign & 0.94 & 0.87 & 0.91  & [20935,3065]\\
\cline{2-5}
& Attack & 0.88 & 0.95 & 0.91 & [1303,22697]\\
\hline
\end{tabular}
\label{table: 9}
\newline
\caption{Federated Model results using our novel min-max scalar and \emph{FedSam}}
\end{table}

\subsection{FedSam vs. FedMMB}
Using our optimized FL model, we compared the effectiveness of our \emph{FedSam} algorithm to the \emph{FedMMB} algorithm. In this experiment, we trained two separate federated models with 3 clients using the CIC-IDS2017, CIC-IDS2018, and NCC-DC datasets. To keep it consistent, each model was trained for 5,000 rounds and used our novel-min max algorithm. \emph{FedSam} and \emph{FedMMB} are both viable algorithms, but the loss curve for \emph{FedSam} in figure 2 highlights its strong ability to handle clients with different dataset sizes and distributions. \emph{FedSam} is able to negate the push/pull effect from clients with different amounts of data and aggregate the model weights in a more effective manner than a basic \emph{FedAvg} algorithm and even \emph{FedMMB}. Model 1 was trained using \emph{FedMMB} and model 2 trained using \emph{FedSam}

Both models were trained using the following clients
 \begin{description}
    \item[1.] Client 1 trained using CIC-IDS2018 data
    \item[2.] Client 2 trained using CIC-IDS2017 data
    \item[3.] Client 3 trained using NCC-DC data
\end{description}

\begin{table}[ht!]
\centering
\renewcommand{\arraystretch}{1.5}
\begin{tabular}{ |m{2.75em}|m{2.75em}| m{2.75em}|m{2.75em}|m{3.5em}|m{5.5em}|} 
\hline
\textbf{Model} &
\textbf{Data} & \textbf{Pre-\newline cision} & \textbf{Recall} & \textbf{F1 \newline Score}  & \textbf{Confusion\newline Matrix}

\\\hline
1 & Benign & 0.97 & 0.95 & 0.96 & [22727,1273]\\
\cline{2-5}
& Attack & 0.95 & 0.97 & 0.96 & [716,23284]\\
\hline
2 & Benign &0.98 & 0.97 & 0.97 & [23183,817]\\
\cline{2-5}
& Attack & 0.97 & 0.98 & 0.97 & [458,23542]\\
\hline
\end{tabular}
\label{table: 10}
\newline
\caption{\emph{FedMMB} FL model compared to \emph{FedSam} FL model}
\end{table}
 
\begin{figure}[ht!]
    \includegraphics[width=8cm]{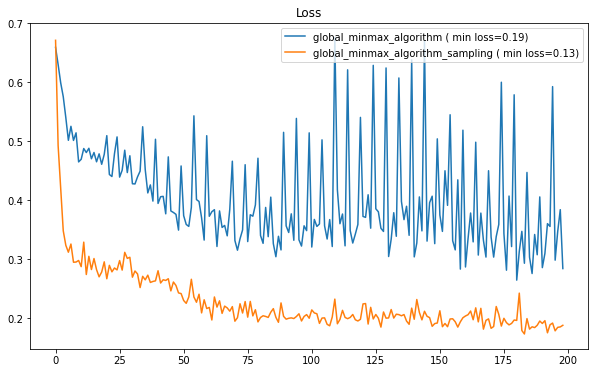}
    \centering
    \caption{\emph{FedMMB} vs \emph{FedSam} Test Data Loss for 2000 Rounds Rounded}
    \label{fig:2}
\end{figure}

\subsection{Unknown Data}

The experiment in this section tested our 3-client FL model with an unknown data set. The results from this experiment highlight our FL model's strong ability to take in unknown data and classify it correctly.

The FL models in this experiment were trained for 5,000 rounds using our novel MinMax Scalar and \emph{FedSam} algorithm. The three clients were trained with either the CIC-IDS2017, CIC-IDS2018, NCC-DC, or the MAWI-Lab datasets and tested with dataset not used in the training. Below are the test datasets used for each model. See table XII for the results. 

 \begin{description}
    \item[1.] Model 1 tested with MAWI-Lab data
    \item[2.] Model 2 tested with CIC-IDS2018 data
    \item[3.] Model 3 tested with CIC-IDS2017 data
    \item[3.] Model 4 tests with NCC-DC data
\end{description}

\begin{table}[ht!]
\centering
\renewcommand{\arraystretch}{1.5}
\begin{tabular}{ |m{2.25em}|m{2.75em}| m{2.75em}|m{2.75em}|m{3.5em}|m{6.75em}|} 
\hline
\textbf{Model} &
\textbf{Data} &\textbf{Pre-\newline cision} & \textbf{Recall} & \textbf{F1 \newline Score} & \textbf{Confusion\newline Matrix}

\\\hline
1 & Benign & 1.0 & 0.89 & 0.94 & [1333523,166477]\\
\cline{2-5}
& Attack & -- & -- & -- & [0,0]\\
\hline
2 & Benign & 1.0 & 0.96 & 0.98  & [2134210,84352]\\
\cline{2-5}
& Attacks & 0.95 & 1.0 & 0.97 & [2365,1621125]\\
\hline
3 & Benign & 0.99 & 0.93 & 0.96 & [1033264,71900]\\
\cline{2-5}
& Attack & 0.88 & 0.99 & 0.93 & [6340,519863]\\
\hline
4 & Benign & 0.71 & 1.0 & 0.83  & [23183,817]\\
\cline{2-5}
& Attack & 1.0 & 0.93 & 0.97 & [28786,403084]\\
\hline
\end{tabular}
\label{table: 11}
\newline
\caption{\emph{FedSam} on Different Unseen Datasets}
\end{table}

\section{Discussion}
The research conducted demonstrates that \emph{FedSam}, our novel min-max algorithm, and classifier technique are valuable strategies that can help companies with different data distributions use FL to expand the scope of their IDS while keeping their data private. By implementing our novel min-max algorithm and \emph{FedSam}, we were able to scale each client equally and prevent one client from having too much influence in the global model. These techniques helped our anomaly detector via FL preform just as well as a central model trained with all of the data sets. Additionally, the individual central models have relatively low accuracies of 69.76\%, 62.68\%, and 41.24\% when tested on a combination of CIC-IDS2018, CIC-IDS2017, and NCC-DC data. If we imagine each of these models as separate companies, their IDS would perform poorly. However, when the three companies participate in the federated model, they each have access to the global IDS that classifies benign and malicious network activity correctly in 98\% of scenarios.  

Another benefit of FL is data privacy and protection. Because our novel min-max algorithm is randomly initialized before being sent to a client, the data used to create the global min-max scaler can never be tracked back to a specific client. This keeps the small amount of data used to create the global min-max scaler anonymous.

For these reasons, our approach to anomaly detection via FL with our novel techniques provide a way for companies of different sizes to share their insights with each other in a secure manner and improve the scope and effectiveness of their IDS.  

\section{Conclusion}
In this paper, we presented a novel min-max scalar for FL, a sampling technique called \emph{FedSam} which helps ensure each client in the federation has an equal influence on the model, and an autoencoder with classifier model trained using FL to detect benign and malicious network activity. Our results showed that our novel FL techniques have the ability to perform very well in the field of anomaly detection, and it can become a valuable asset to a company's IDS.  
\subsection{Future Work}
There are a variety of topics to continue our research in FL. One topic is to explore ways to detect when a client, either intentionally or unintentionally, starts using attack data to train the autoencoder. This scenario would cause the anomaly detector to miss attacks and leave the clients in the federation vulnerable. Along with detecting data poisoning, future research could explore ways to detect model poisoning. It is important to ensure the reliability of the global FL model so there needs to be a way to detect when a client has tampered with the model parameters to disrupt the learning.

\bibliographystyle{IEEEtran}
\bibliography{ref}

\end{document}